Ari Goodman, Ryan O'Shea, Noam Hirschorn, Hubert Chrostowski
Naval Air Warfare Center Aircraft Division Lakehurst

# Assurance for Deployed Continual Learning Systems

## Abstract

The future success of the Navy will depend, in part, on artificial intelligence. In practice, many artificially intelligent algorithms, and in particular deep learning models, rely on continual learning to maintain performance in dynamic environments. The software requires adaptation to maintain its initial level of performance in unseen situations. However, if not monitored properly, continual learning may lead to several issues including catastrophic forgetting in which a trained model forgets previously learned tasks when being retrained on new data. The authors created a new framework for safely performing continual learning with the goal of pairing this safety framework with a deep learning computer vision algorithm to allow for safe and high-performing automatic deck tracking on carriers and amphibious assault ships. The safety framework includes several features, such as an ensemble of convolutional neural networks to perform image classification, a manager to record confidences and determine the best answer from the ensemble, a model of the environment to predict when the system may fail to meet minimum performance metrics, a performance monitor to log system and domain performance and check against requirements, and a retraining component to update the ensemble and manager to maintain performance. The authors validated the proposed method using extensive simulation studies based on dynamic image classification. The authors showed the safety framework could probabilistically detect out of distribution data. The results also show the framework can detect when the system is no longer performing safely and can significantly extend the image classifier's working envelope. Future follow-on work will include increasing the fidelity of the images to mimic views from the Integrated Launch and Recovery Television Surveillance/System currently installed on carriers, as well as experiments focused on prediction and certification of the overall system.

## Introduction

There is an inherent need for strong capabilities in accurately monitoring, controlling, and certifying learning systems because such learning systems will be used in military contexts where safety and reliability are paramount. Improper monitoring can lead to several issues, such as catastrophic forgetting, which is the tendency of a learning system to forget previously learned information when learning new information [1].

The objective of this work is the development of an architecture capable of monitoring, adaptation, and certification. Monitoring is critical to accurately determine when the underlying system fails to meet requirements. Adaptation is critical to allow the underlying system to change depending on stimuli from the deployed environment. The deployed environment is necessarily different from the trained environment, and these differences provide the opportunity for the underlying system's performance to be improved through online retraining. Certification is critical to ensure that the system will continue to meet requirements in the future and to define the requirements for continuous recertifications and inspections.

In this paper, experimental evidence is provided to support the proposed framework's inherent safety and learning properties with respect to the monitoring and retraining components.

## Background

The standard paradigm in machine learning is to train a system on an initial dataset, called a training dataset, and once the system achieves appropriate performance on a holdout dataset, called a validation or testing dataset, save the model and end modifying the network weights. Continual learning, also called life-long learning and incremental learning, differs from this paradigm by allowing the network weights to continue to be modified after saving the model by learning from new, incoming data. Although the incoming data is unlabeled, the data can be clustered using unsupervised learning techniques to either form new classes or update existing classes [2-4].

Continual learning has the advantage of improving adaptability to changing domains and fixing errors in the initial sampling that formed the training dataset. However, continual learning also allows for the possibility of mislabeling incoming data, therefore reducing overall performance. Also, chronic forgetting, when the model performs poorly on already trained tasks, is a key risk when modifying network weights [2-4].

There are techniques to increase the underlying safety for deep learning and in particular continual learning. For example, formal methods like control barrier functions can be used to guarantee that a learned controller will not violate specifications [5-7]. Runtime monitors and verifiers can be used to guarantee if any specification is violated, the system would default to a safety behavior [8]. Systems can include human-in-the-loop interfaces for people to oversee any changes that the learning system makes [9]. Certification techniques for machine learning systems exist to ensure safety and guarantee performance [10-12]. There are also existing works specifically studying the safety of deep learning in a continual learning context [13].

## Methodology
### Framework Components
The following general framework is proposed:

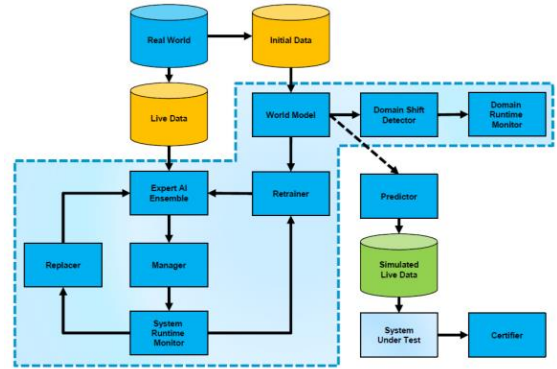

Figure 1: Proposed Safety Framework Diagram

The highlighted blue section in Figure 1 is the focus of this paper. The components in this framework are meant to be replaceable by new state-of-the-art architectures.

First, there are three models of sources of information from the real-world: Real World, Initial Data, and Live Data Stream. The Real World represents all possible classes, disturbances, augmentations, etc. for the data for all time. The Initial Data is a subset of data collected over a certain period from the Real World, possibly with other constraints such as certain classes, augmentations, disturbances, etc. The Initial Dataset is used to train and test all components before runtime. The Live Data Stream is a subset of the Real World that is collected over a continuous period. This data is used during runtime. The distinct data sources, although similar, have natural fluctuations due to random sampling built in, which is meant to test the robustness and adaptability of the architecture.

Next, the World Model is a model used for synthesizing data which is similar to the Real World data. The World Model is trained on the Initial Dataset. In several experiments, an autoencoder was used to represent the world model, but more state-of-the-art architectures may be appropriate for more realistic scenarios. To isolate error in some experiments, the World Model was replaced directly by the Real World. By using this replacement, the error with respect to accurately mimicking data from the Real World are isolated.

An Expert AI Ensemble receives Live Data and produces an answer and an array of trust scores. Trust scores are an array of values from various metrics. In the experiments, the following metrics were used:

1. An autoencoder trained on the Expert AI's training dataset's reconstruction loss
2. Earth's Movers Distance [15] between the incoming image and the closest neighbor in the Expert AI's training dataset
3. Binary value if k-nearest-neighbors on the closest neighbors in the Expert AI's training dataset matches the Expert AI's prediction on the incoming image [16]
4. The difference between the top two Softmax values
5. The variation ratio of the Softmax values
6. The entropy amongst the Softmax values
7. The Softmax value for each class

It is possible to only have one AI in the ensemble or heterogeneous AI architectures, but for the experiments in this paper there were several experts of the same architecture trained on different Initial Data sets.

The Expert AI Ensemble was made of shallow convolutional neural networks with two sets of weights and biases each, a semi-static, or reserved, set of weights and biases and a dynamic, or active, set of weights and biases. The semi-static set of weights and biases would be routinely replaced by the dynamic set of weights and biases if the following conditions were met:

1. Human supervisor approval was given, or the human approval required flag was set to False
2. The performance of the new sets of weights met safety and performance standards
3. The performance of the current sets of weights did not meet safety or performance standards; or the performance of the current sets of weights was worse than the new sets of weights, and performance maximizing flag was set to True

The Domain Shift Detector detects deviations in the Live Data Stream and discrepancies between the World Model and Live Data Stream. This information is provided to the Domain Runtime Monitor which logs any deviations and verifies that the system under test is operating within its working envelope. In this work, the Domain Shift Detector and Domain Runtime Monitor were represented by autoencoder architectures. These autoencoders were trained on the Initial Dataset, and the error between their reproduced images and the incoming images was used to determine in or out of domain data. Similarly, each Expert AI, under metric 1, was paired with a domain monitor to determine the similarity between the Expert AI's training data and incoming data.

The System Run Time Monitor ensures the components are performing within their specified performance requirements. The System Run Time Monitor receives information from the Manager with respect to the Expert AI's trust metrics. It is constantly checking against requirements to ensure the current state of the system is not violating the operational specification. The Runtime Monitor includes both a set of specifications to check the metrics against, as well as a testing dataset to ensure the system under test is not experiencing chronic forgetting. More than just simply shutting down immediately, several reactions are built into the System Run Time Monitor depending on the situation. The System Run Time Monitor can send out notifications to users when specifications are violated. It can also signal the Retrainer and Replacer to initiate retraining. Finally, it can shut down the system entirely. It should be noted that accuracy on Live Data cannot be directly measured and therefore cannot be used for the System Run Time Monitor, but an experiment was run to determine the correlation between the trust metrics and accuracy, and to use trust metrics as a proxy for Live Data accuracy.

The Manager receives information from the Expert AI Ensemble and determines the final response. The Manager is trained using the Expert AI Ensemble's known performance and trust metrics from the Initial Dataset. It was necessary to include training data in the Initial Dataset when none of the Expert AIs should be trusted to train the Manager. The Manager is based on a CNN architecture and weights the responses from the Expert AIs according to their trust metrics. If the Manager determines none of the existing Expert AIs meet the trust tolerance requirement, the Manager will signal the Retrainer and Replacer to train a new Expert AI or to retrain an existing Expert AI.

The Retrainer and Replacer use incoming data to update the Expert AIs and their datasets. These components are necessary to perform continual learning, but the methods to produce a Retrainer and Replacer here within are experimental. When the aforementioned three conditions for the Expert AI Ensemble retraining are met, the Retrainer and Replacer can replace the semi-static weights with the dynamic versions. During runtime, incoming uncertain data is flagged by the Manager and clustered by the Retrainer. Meanwhile, confidently labeled data is used by the Replacer to modify the datasets of the AI's. The Retrainer and Replacer have tunable parameters representing how quickly to make new classes, overwrite old data, and distribute new training.

The Predictor uses the World Model to create data that is likely to occur in the future and forward simulate the performance of all system components.

The Certifier verifies that the predicted behavior of the system does not violate its operating specifications. The Certifier uses predicted data to determine if the system will perform as expected for a certain time horizon; it also determines when and how likely the system is to fail.

**Dataset**

The dataset used for the experiment was a set of grayscale 28x28x1 tensor images of digits rotated from 0 to 360 degrees similar to Rotated MNIST [14]. Varying amounts of gaussian noise were added to the images. Subsets of this dataset were taken based on the digit and the angle to represent different domains, e.g., digits rotated between 0 and 30 degrees or 3's and 4's. An example set of digits is shown in Figure 2 below.

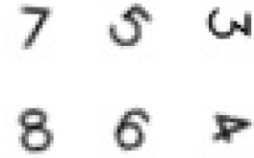

Figure 2: Dataset of Rotated Digits

**Experiments**

**World Model and Domain Shift Detector Experiment**

The Domain Shift Detector component's ability to accurately detect in and out of domain data from a pretrained World Model was quantified by training several autoencoders on different Initial Datasets and measuring the reconstruction loss on the Real World dataset. The autoencoders were used as the World Model Components and their Reconstruction Loss served as the Domain Shift Detection. Below in Figure 3 is data from the World Model trained on 0-10 degrees of rotation on all 10 digits when tested on all 10 digits rotated between 0 and 100 degrees. Only the 0–10-degree rotation was in domain.

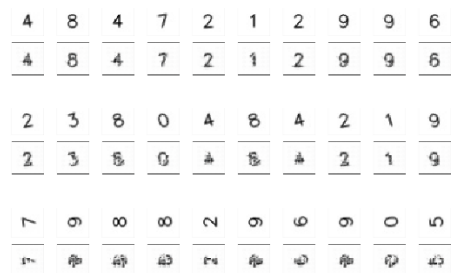

Figure 3: Reconstructed Digits in the Range 0-10, 10-20, and 90-100 Degrees of Rotation

Four sets of trials were conducted where the World Model component reconstructed incoming images over the range 0-100 degrees of rotation. Each set of trials were run 10 times and the results are averaged. In the first set of trials, the World Model was trained on 0-10 degrees of rotation. In the second set of trials, the World Model was trained on 0-100 degrees of rotation. In the third set of trials, it was trained on 0-50 degrees of rotation. In the fourth set of trials, it was trained on 40-50 degrees.

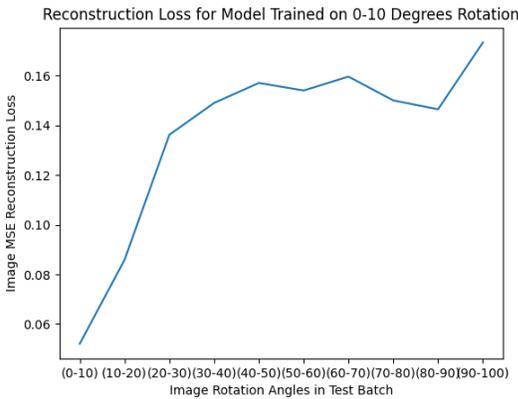

Figure 4: World Model Trained on 0-10 Degrees

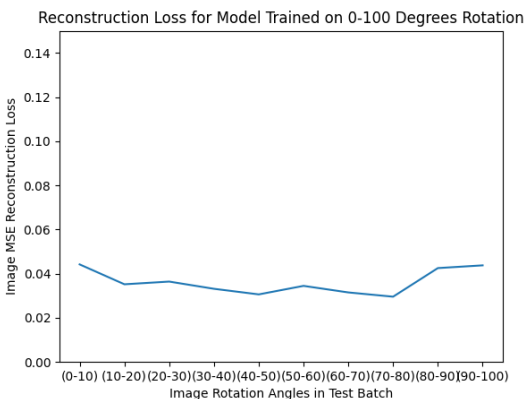

Figure 5: World Model Trained on 0-100 Degrees

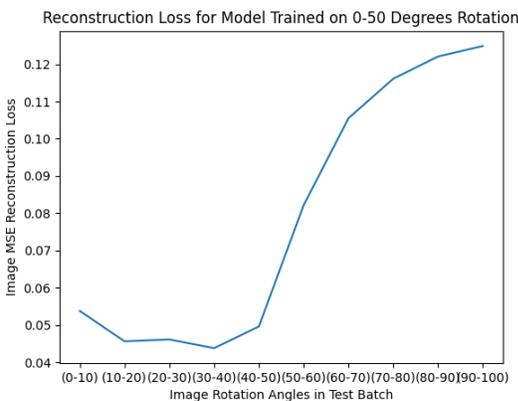

Figure 6: World Model Trained on 0-50 Degrees

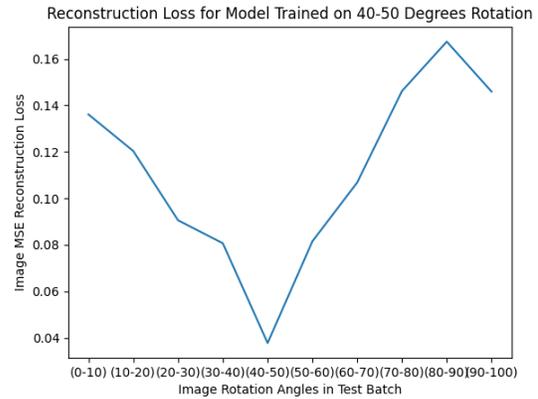

Figure 7: World Model Trained on 40-50 Degrees

Figures 4-7 show a correlation between the reconstruction loss for a given autoencoder and if the incoming data was in domain or out of domain. Also, there is a correlation between the reconstruction loss and the degree to which the incoming data was out of domain.

**Expert AI Experiment**

One Expert AI with a CNN architecture was trained on an initial dataset with images rotated between 0-10 degrees. This AI was subsequently tested with samples having a rotation between 10 and 20 degrees and between 90 and 120 degrees.

To train the Expert AI, small batch and training sets were used (80 images training, 20 images validation, batch size 10), and it was tested with 1,000 images. The loss function used was cross entropy. The optimization function used was Stochastic Gradient Descent. The learning rate chosen was 0.01 and moment 0.5.

The Expert AI was able to achieve optimal performance, 100% accuracy, on out of distribution data (10-20) that was similar to the training set (0-10) as shown in Figure 8 below. As shown in Figure 9 below, it performed poorly, 22% accuracy, on significantly out of distribution data (90-120), and would therefore need an adaptation component to meet specifications.

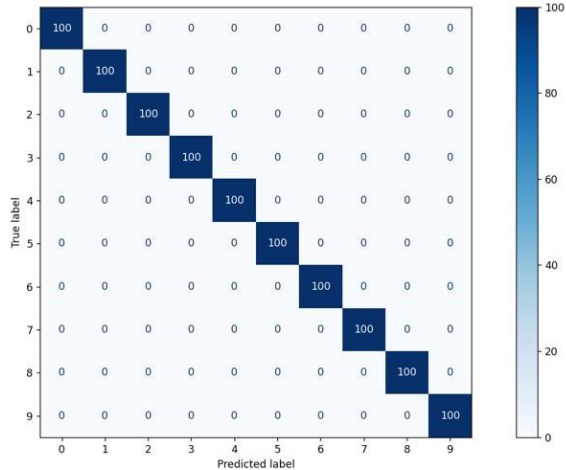

Figure 8: Confusion Matrix on Slightly Out of Distribution Data (10-20 degrees)

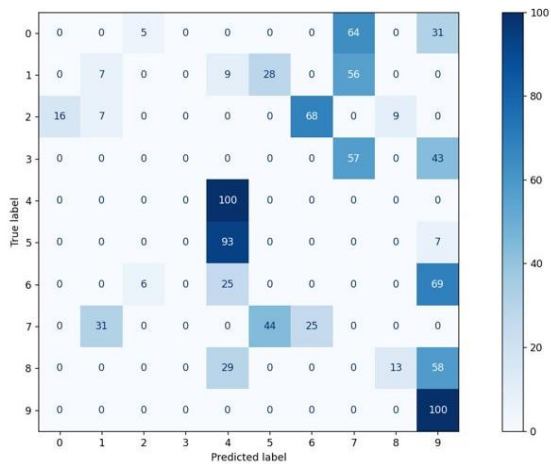

Figure 9: Confusion Matrix on Significantly Out of Distribution Data (90-120 degrees)

**Retrainer, Replacer, Manager, and Run Time Monitor Experiment**

The Retrainer and Replacer component's abilities to adapt the Expert AI Ensemble to out of distribution data was quantified in a series of trials. The correlation between accuracy and confidence as reported by the Manager as reported by the Run Time Monitor was quantified as well.

In both sets of trials, a Manager and an Expert AI Ensemble was trained on an initial dataset from 0-20 degrees of rotation. A System Runtime Monitor was used to extract the performance metrics from the Expert AI Ensemble. Replacer and Retrainer components were integrated to allow the Expert AI Ensemble to adapt to the changing Live Data. Each set of trials were run three times and the results averaged.

In the first set of trials, the incoming data was changed at 10 degrees per timestep and included a system with a Retrainer and a fast Replacer, a Retrainer and a slow Replacer, no Retrainer and a slow Replacer, no Retrainer and a fast Replacer, and neither a Retrainer nor a Replacer.

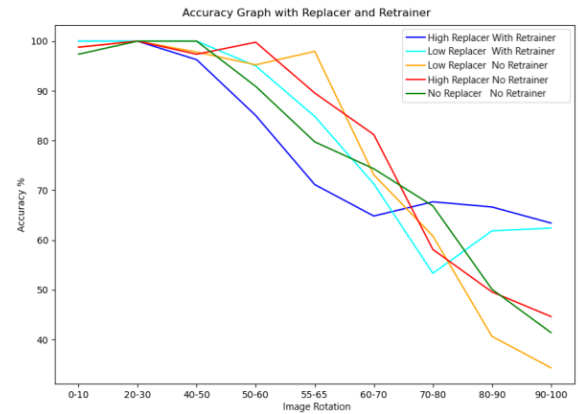

Figure 10: Accuracy of Components on Quickly Changing Data

In this first set of trials, the Retrainer and Replacer were not able to adapt the Expert AI Ensemble adequately to the incoming data. As can be seen in Figure 10, the Retrainers and Replacers did not significantly impact the performance of the system.

In the second set of trials, the incoming data was sampled from different ranges, increasing by 5-20 degrees per timestep and included a system with a Retrainer and a fast Replacer, a Retrainer and a slow Replacer, and neither a Retrainer nor a Replacer.

Figure 11 below shows that unlike the first set of trials, the Retrainer and Replacer were able to adapt the Expert AI Ensemble adequately. The accuracy with a Retrainer and a high Replacers did improve the performance of the system and extend the useability significantly outside its originally trained dataset. The Run Time Monitor alerted failed performance specifications in the 50-60 timestep block based on the low-confidence of the system for the Retrainer with Low Replacer and no Retrainer or Replacer.

The confidence values shown in Figure 11 represent the maximum trust value from any individual Expert AI averaged over the image interval. There is a correlation between the confidence values reported by the Manager and the performance of the Expert AI Ensemble.

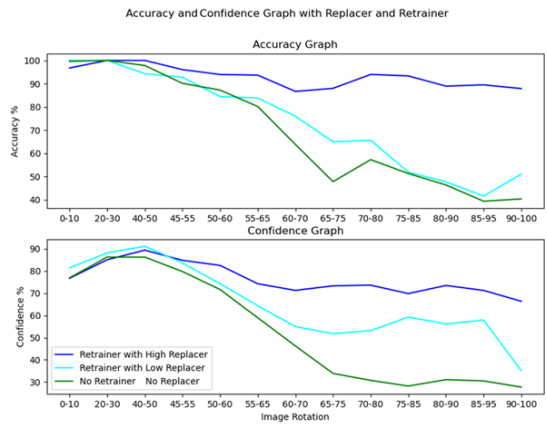

Figure 11: Accuracy and Confidence of Components on Slowly Changing Data

## Conclusion

In this work, a prototype framework was presented with the ability to monitor, adapt, and certify an underlying computer-vision classification machine learning model.

Through a series of experiments, it was shown that the underlying system's performance was able to be monitored accurately using the Domain Shift Detector component. Even without labels or unsupervised clustering, it is possible for the Domain Runtime Monitor to use the information from the Domain Shift Detector to determine if the incoming data is within the system's working envelope or training dataset.

The Retrainer and Replacer components were shown to introduce an appropriate amount of adaptation for the underlying system under certain conditions. Experiments showed the system with the Retrainer and Replacer can maintain a higher level of accuracy than a system without these components. The experiments also showed the limits of the framework when compared to optimal performance in unfavorable scenarios.

The architecture was successful in extending the safe working envelope and accurately monitoring behavior and performance.

In the future, more rigorous experiments and a more difficult dataset will be used to qualify the generality of the framework to more realistic scenarios. Additional experiments will be performed to evaluate its prediction and certification capabilities.

## References

1. Goodfellow, I. J., Mirza, M., Xiao, D., Courville, A., & Bengio, Y. (2013). An empirical investigation of catastrophic forgetting in gradient-based neural networks. *arXiv preprint arXiv:1312.6211*.
2. Schlimmer, J. C., & Granger, R. H. (1986). Incremental learning from noisy data. *Machine learning*, *1*(3), 317-354.
3. Sutton, R. S., & Whitehead, S. D. (2014, May). Online learning with random representations. In *Proceedings of the Tenth International Conference on Machine Learning* (pp. 314-321).
4. Ring, M. B. (1998). CHILD: A first step towards continual learning. In *Learning to learn* (pp. 261-292). Springer, Boston, MA.
5. Gillula, J. H., & Tomlin, C. J. (2012, May). Guaranteed safe online learning via reachability: tracking a ground target using a quadrotor. In *2012 IEEE International Conference on Robotics and Automation* (pp. 2723-2730). IEEE.
6. Srinivasan, M., Dabholkar, A., Coogan, S., & Vela, P. A. (2020, October). Synthesis of control barrier functions using a supervised machine learning approach. In *2020 IEEE/RSJ International Conference on Intelligent Robots and Systems (IROS)* (pp. 7139-7145). IEEE.
7. Akametalu, A. K., Fisac, J. F., Gillula, J. H., Kaynama, S., Zeilinger, M. N., & Tomlin, C. J. (2014, December). Reachability-based

safe learning with Gaussian processes. In *53rd IEEE Conference on Decision and Control* (pp. 1424-1431). IEEE.
8. Kwiatkowska, M., Norman, G., & Parker, D. (2011, July). PRISM 4.0: Verification of probabilistic real-time systems. In *International conference on computer aided verification* (pp. 585-591). Springer, Berlin, Heidelberg.
9. Xin, D., Ma, L., Liu, J., Macke, S., Song, S., & Parameswaran, A. (2018, June). Accelerating human-in-the-loop machine learning: Challenges and opportunities. In *Proceedings of the second workshop on data management for end-to-end machine learning* (pp. 1-4).
10. Winter, P. M., Eder, S., Weissenböck, J., Schwald, C., Doms, T., Vogt, T., ... & Nessler, B. (2021). Trusted Artificial Intelligence: Towards Certification of Machine Learning Applications. *arXiv preprint arXiv:2103.16910*.
11. Dvijotham, K., Gowal, S., Stanforth, R., Arandjelovic, R., O'Donoghue, B., Uesato, J., & Kohli, P. (2018). Training verified learners with learned verifiers. *arXiv preprint arXiv:1805.10265*.
12. Salman, H., Yang, G., Zhang, H., Hsieh, C. J., & Zhang, P. (2019). A convex relaxation barrier to tight robustness verification of neural networks. *Advances in Neural Information Processing Systems*, *32*.
13. Bennani, M. A., Doan, T., & Sugiyama, M. (2020). Generalisation guarantees for continual learning with orthogonal gradient descent. *arXiv preprint arXiv:2006.11942*.
14. Larochelle, H., Erhan, D., Courville, A., Bergstra, J., & Bengio, Y. (2007, June). An empirical evaluation of deep architectures on problems with many factors of variation. In *Proceedings of the 24th international conference on Machine learning* (pp. 473-480).
15. Kusner, M., Sun, Y., Kolkin, N., & Weinberger, K. (2015, June). From word embeddings to document distances.
16. Papernot, N., & McDaniel, P. (2018). Deep k-nearest neighbors: Towards confident, interpretable and robust deep learning. *arXiv preprint arXiv:1803.04765*. In *International conference on machine learning* (pp. 957-966). PMLR.

Ari Goodman is the S&T AI Lead and a Robotics Engineer in the Robotics and Intelligent Systems Engineering (RISE) lab at Naval Air Warfare Center Aircraft Division (NAWCAD) Lakehurst. In this role he leads efforts in Machine Learning, Computer Vision, and Verification & Validation of Autonomous Systems. He received his MS in Robotics Engineering from Worcester Polytechnic Institute in 2017.

Ryan O'Shea is a Computer Engineer in the Robotics and Intelligent Systems Engineering (RISE) lab at Naval Air Warfare Center Aircraft Division (NAWCAD) Lakehurst. His current work is focused on applying computer vision, machine learning, and robotics to various areas of the fleet in order to augment sailor capabilities and increase overall operational efficiency. He received a Bachelor's Degree in Computer Engineering from Stevens Institute of Technology.

Noam Hirschorn is a senior computer engineering student at Rutgers University and was an intern with the Naval Air Warfare Center Aircraft Division (NAWCAD) – Lakehurst in 2022. His research interests include deep learning, specifically with FPGAs, and he will be pursuing potential applications of this technology at graduate school.

Hubert Chrostowski is a mechanical design engineer with the Naval Air Warfare Center Aircraft Division (NAWCAD) – Lakehurst. In his current role, he designs and aids in the acquisition of support equipment for the CH-53K heavy lift helicopter. He received his Bachelor's Degree in Mechanical Engineering from Rutgers University in 2017.